\documentclass[article]{article}
\usepackage{iclr2016_conference,times}
\usepackage{natbib}
\usepackage{amsmath,amsthm,bm,mathrsfs,mathtools,amssymb}
\usepackage{graphicx}
\usepackage{hyperref}
\usepackage{color}

\iclrfinalcopy

\begin{document}

\title{Predicting distributions\\ with Linearizing Belief Networks}

\author{
Yann N. Dauphin, David Grangier \\
Facebook AI Research\\
1 Hacker Way\\
Menlo Park, CA 94025, USA \\
\texttt{\{ynd,grangier\}@fb.com}
}

\date{\today}
\maketitle

\begin{abstract}

Conditional belief networks introduce stochastic binary variables in neural networks. Contrary to a classical neural network, a belief network can predict more than the expected value of the output $Y$ given the input $X$. It can predict a distribution of outputs $Y$ which is useful when an input can admit multiple outputs whose average is not necessarily a valid answer. Such networks are particularly relevant to inverse problems such as image prediction for denoising, or text to speech. However, traditional sigmoid belief networks are hard to train and are not suited to continuous problems. This work introduces a new family of networks called linearizing belief nets or LBNs. A LBN decomposes into a deep linear network where each linear unit can be turned on or off by non-deterministic binary latent units. It is a universal approximator of real-valued conditional distributions and can be trained using gradient descent. Moreover, the linear pathways efficiently propagate continuous information and they act as multiplicative skip-connections that help optimization by removing gradient diffusion. This yields a model which trains efficiently and improves the state-of-the-art on image denoising and facial expression generation with the Toronto faces dataset.

\end{abstract}

\section{Introduction}
\begin{figure}[t]\label{fig:unimodal}
\centering
\includegraphics[width=0.6\textwidth]{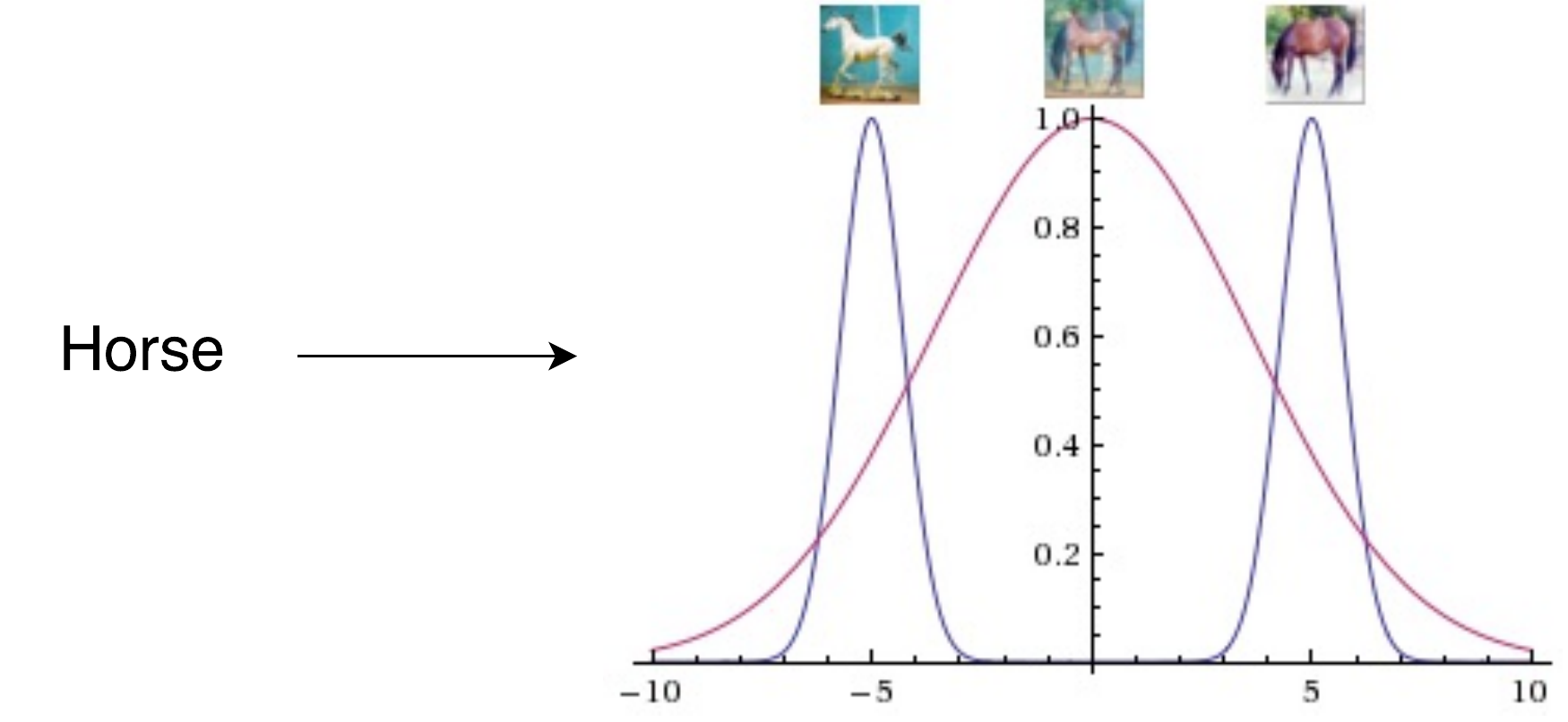}
\caption{The optimal setting of a uni-modal Gaussian (purple) for a distribution with just two modes (blue) results in an incorrect density model and a high standard deviation. This is a simple illustration of the averaging of outcomes we observe in practice for more interesting problems.}
\end{figure}

Deep neural networks are universal approximators that can learn any deterministic mapping $f: X \to Y$ given enough capacity. However, traditional neural networks are not universal approximators of conditional distributions $p(Y|X)$. In the context of continuous data, neural networks with the mean squared error can be derived from maximum likelihood on a unimodal Gaussian distribution  $p(Y|X) = \mathcal{N}(\mu=f(X), \sigma=1)$ where the network $f$ predicts the expected value. Thus conventional networks could not learn output distributions with multiple modes. This kind of distribution occurs for instance when trying to predict an image $Y$ based on a description $X$. This distribution would be the set of images that fits the description - not a single image. In general, a similar situation occurs for most ill-posed or inverse problems - whenever the model does not have enough information to rule out all uncertainty over outcomes. In these situations, the unimodal prior forces the network to average the outcomes as illustrated in Figure \ref{fig:unimodal}. This is problematic because in many cases this generates an invalid prediction and in the case of images this is exemplified by blurry average predictions. We observe that this occurs in several important applications of neural networks to the continuous domain -- i.e. predicting the next frame of video \citep{srivastava2015unsupervised} or learning unsupervised models with regularized autoencoders \citep{bengio2013deep}.

Stochastic feed-forward neural networks \citep{neal1992connectionist} (SFNN) solve this problem with the introduction of stochastic latent variables to the network. The model can be seen as a mixture of neural networks where each configuration of stochastic variables defines a different neural network. This is efficiently achieved by sharing most of the parameters between configurations. While conventional neural networks fit a single conditional Gaussian to the data, the stochastic latent variables lead to fitting a mixture of conditional Gaussians. This a powerful extension since mixture of Gaussians are universal approximators of distributions \citep{Sorenson:1971}. The network can model multi-modal distributions by learning a different network for each mode. \cite{neal1992connectionist} proposes training Sigmoid Belief Networks (SBN) which have only binary stochastic units. The resulting model makes piecewise-constant MAP predictions and thus is unsuitable for continuous problems -- it cannot vary smoothly with the input, see Section~\ref{sec:related_work}. \cite{tang2013learning} addresses this limitation with the addition of deterministic sigmoid units to each layer of the network. This yields a mixture of non-linear neural networks gated by a stochastic non-linear neural network. \cite{tang2013learning} showed improved results with this model but the training of the latent stochastic units with a high variance variational bound was a challenge. \cite{raiko2014techniques} suggested to avoid training the latent units, relying only on layers of deterministic units to shape the random distribution. This modification did not however eliminate a fundamental limitation of stochastic networks in which stochastic and deterministic units interact additively. In these networks, the gradients of the weights tied to deterministic units have much lower variance than those tied to stochastic units, which means it is harder assign credit to stochastic units and training prefers configuration using the deterministic ones as much as possible.


In this paper, we propose a new class of stochastic networks called linearizing belief nets (LBN) that can learn any continuous distribution $p(Y|X)$. This model combines deterministic variables with non-deterministic binary variables in a multiplicative fashion. This approach allows using linear deterministic units without losing modeling power. These linear units can be thought of as multiplicative skip connections that allows the gradient to flow without diffusion through deep networks~\citep{hochreiter2001}. Furthermore, multiplicative interactions allow making tie-breaking choices which would be difficult to emulate with addition. Our experiments on facial expressions confirm that the model can successfully learn multimodal distributions. We demonstrate with image denoising that the model can attain state-of-the-art results modeling natural images - converging faster than even deterministic ReLU networks in some cases.

\section{Linearizing Belief Nets}

\begin{figure}[h]\label{fig:piecewise}
\centering
\includegraphics[width=0.7\textwidth]{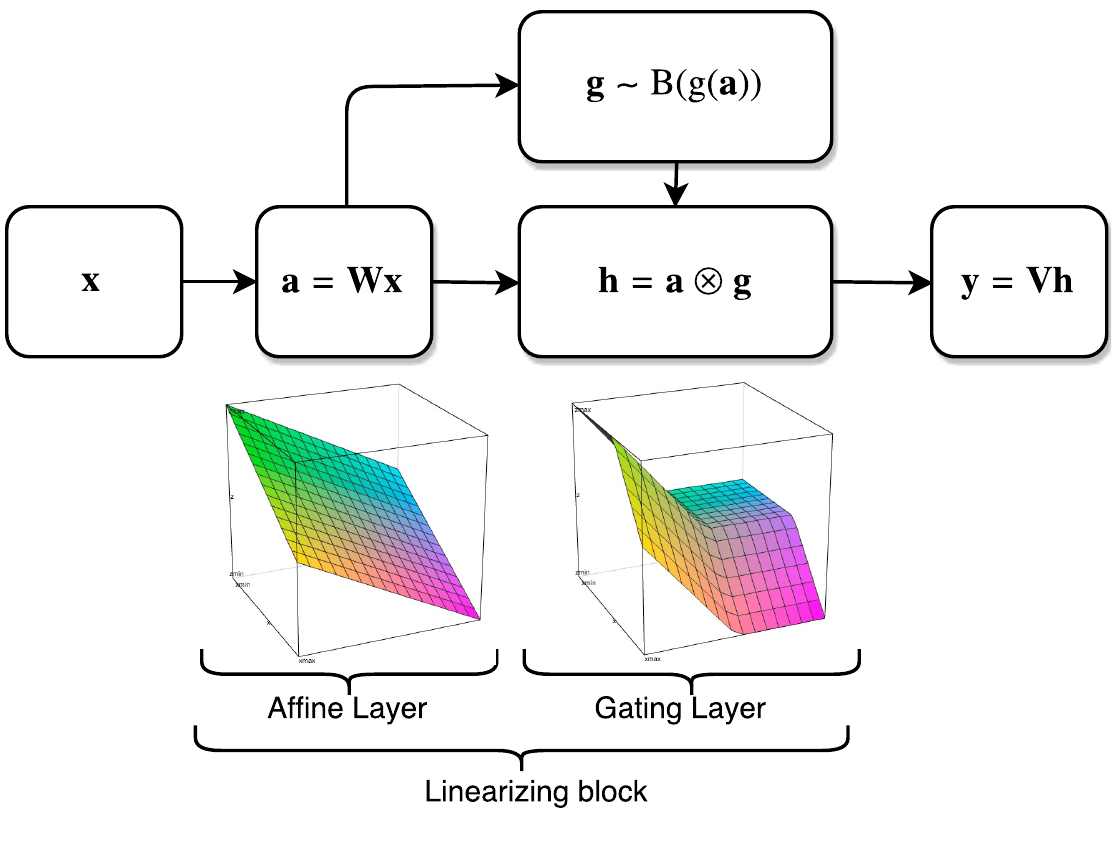}
\vspace{-6mm}
\caption{Flow chart of a LBN with a single gating layer. The information flows linearly through the active units and the gating units introduce non-linearity by deactivating units. The charts at the bottom illustrate how each block can transform its input in a 2D case.}
\end{figure}

This section introduces a new family of belief networks dubbed linearizing belief nets (LBN) aimed to address some of the limitations of conditional SBNs. 
These models factor into a deep linear network where each linear unit is gated by a non-deterministic binary unit. It can be seen as a non-linear mixture of an exponential amount of linear models and it can learn piece-wise linear non-deterministic functions as illustrated in Figure \ref{fig:piecewise}. The binary gating units can select the appropriate specialist sub-network for an input by turning off linear units. For instance, the mixture could contain different specialist sub-networks to generate different types of dogs. Let us consider the problem of predicting the target ${\bf y} \in \mathbb{R}^M$ from the input ${\bf x} \in \mathbb{R}^N$. In general, a conditional mixture model can be written as
\begin{equation}\label{eqn:mixture}
p({\bf y}|{\bf x}) = \sum_{\bf g} p({\bf y},{\bf g}|{\bf x}) =  \sum_{\bf g} p({\bf g}|{\bf x}) p({\bf y}|{\bf x},{\bf g})
\end{equation}
where $p({\bf g}|{\bf x})$ is the probability of selecting the expert identified by the gating configuration ${\bf g}$ and $p({\bf y}|{\bf x},{\bf g})$ is the prediction of that expert. Contrary to classical mixtures like Gaussian mixture models (GMM), stochastic networks share parameters across experts which enable training exponentially more experts.

The output of the network $p({\bf y}|{\bf x},{\bf g}) = \mathcal{N}(\mu=f({\bf x}, {\bf g}))$ for a given input and gating configuration is
\begin{align}
{\bf h} & = {\bf g}\otimes{\bf W}{\bf x}\\
f({\bf x}, {\bf g}) & = {\bf V}{\bf h} + {\bf b}
\end{align}
where $\otimes$ denotes component-wise vector product, ${\bf g} \in \{0, 1\}^H$ are the binary gating units and ${\bf V} \in \mathrm{R}^{M\times H}, {\bf W} \in \mathrm{R}^{H\times N}, {\bf b} \in \mathrm{R}^M$ are learned parameters. Each expert is a particular combination of a multi-layer linear network. The non-linearity of the model comes from the gating units ${\bf g}$ turning linear factors on or off -- much like a rectifying non-linearity ${\text{ReLU}({\bf W}{\bf x}) = \text{max}(0, {\bf W}{\bf x}) = ({\bf W}{\bf x} > 0)\otimes {\bf W}{\bf x}}$. We show that linearizing blocks generalize the ReLU in Appendix \ref{sec:relu}. The binary units can be seen as controlling the flow of information in the network. At the same time, the linear factors transmit continuous values, which is crucial to address continuous prediction tasks. Hence the model disentangles the problem of deciding when to fire and how much to fire. This model is one of the simplest ways to learn smooth functions with binary latent variables. The set of experts are the $2^H$ different configurations of the binary units ${\bf g}$. Now a key question is how to learn an appropriate distribution over these units efficiently.

The gating units are sampled according to a Bernoulli distribution $p({\bf g}|{\bf x}) = \mathcal{B}(p=g({\bf W}{\bf x}))$, where each unit ${\bf g}_i$ is sampled independently with rate $g_i({\bf W}{\bf x})$ set by the non-linear gating function $g$. Section~\ref{sec:implementation} discusses the specific parameterization of $g$. The gating network decides which units to activate/deactivate and allows modeling complex patterns. Therefore, the gating function takes the linear activation vector ${\bf W}{\bf x}$ as input and can implement rich interactions between factors such as winner-takes-all. Adding such pooling interactions to the networks is known to improve the generalization performance \citep{boureau2010theoretical,goodfellow2013maxout}. We propose these non-linear gating units as a general strategy to learn pooling functions. 
In addition, the gating units allows the model to easily represent sparse activations. This is an interesting property since the reconstruction of high frequency signals by summing a few high frequency dictionary elements is a common, effective strategy~\citep{mallat:2008}. In our experiments, we oberve that our model indeed learns sparse, orthogonal features, see Section~\ref{sec:experiments}.


The conditionals corresponding to $f$ and $g$ defines the model. Different objectives can be employed to train their parameters: data likelihood, variational bounds~\citep{tang2013learning}, or a distiguishibility criteria~\citep{goodfellow2014generative}. This work focuses on maximum likelihood and we estimate the computationally expensive expectation from Equation \ref{eqn:mixture} using Monte Carlo with $k$ samples
\begin{equation}
\log p({\bf y}|{\bf x}) \simeq \log \frac{1}{k} \sum_{i=1}^k \frac{1}{\sqrt{(2\pi)^M}} ~ \exp\left( -\frac{1}{2}\|f({\bf x}, {\bf g}^{\{i\}}) - {\bf y}\|^2 \right)
\end{equation}
where ${\bf g}^{\{1\}}, \ldots, {\bf g}^{\{k\}}$ are $k$ samples from 
$p({\bf g}|{\bf x})$.

The model $p({\bf y}|{\bf x})$ is a mixture of Gaussians and can therefore approximate any conditional distribution~\citep{Sorenson:1971}. In contrast, traditional neural networks estimate the parameters of a single Gaussian.
Differentiating the loss 
reveals the presence of a softmax between the Gaussians
\begin{equation}
-\frac{\partial}{\partial \theta}\log p({\bf y}|{\bf x}) 
\propto 
\sum_{i=1}^k 
\text{softmax}\left(-\frac{1}{2}\|f({\bf x}, {\bf g}^{(i)}) - {\bf y}\|^2\right) \frac{\partial}{\partial \theta} 
\left\|f({\bf x}, {\bf g}^{(i)}) - {\bf y}\right\|^2.
\end{equation}
The gradient concentrates on the sample $f({\bf x}, {\bf g}^{\{i\}})$ closest to  target $y$ and updates the model to move $f({\bf x}, {\bf g}^{\{i\}})$ toward target ${\bf y}$. Thus the model can learn different Gaussians to account for the diverse set of alternative outcomes ${\bf y}$ for the input ${\bf x}$. One can further note that relying on a single Monte Carlo sample ($k=1$) reverts to mean square error minimization. The combination of the linear skip-connections with binary latent variable helps learning as it prevents gradient diffusion~\citep{hochreiter2001}. For instance, in the gradient
$$
\frac{\partial \log p({\bf y}|{\bf x})}{\partial {\bf W}_{i,j}} 
= 
{\bf g}_j
~
{\bf x}_i
~
\frac{\partial \log p({\bf y}|{\bf x})}{\partial {\bf h}}  + \frac{\partial {\bf g}}{\partial {\bf W}_{i,j}}\frac{\partial \log p({\bf y}|{\bf x})}{\partial {\bf h}}
$$
there is a path for the gradient to flow without down weighting through the network. ${\bf g}_j \in \{0,1\}$  either selects the full gradient or cancels it much like in a ReLU networks.

\section{Learning non-deterministic gating units}
\label{sec:implementation}

The rates of the gating units ${\bf g} \sim \mathcal{B}(p=g({\bf W}{\bf x}))$ are parameterized by the gating function $g$. We implement $g$ with a neural network which keeps training simple without compromising the power of the model. A simple choice would be to use $g({\bf W}{\bf x}) = \sigma({\bf W}{\bf x} + {\bf b})$ with the sigmoid function ${\sigma: x \to \frac{1}{1+\exp(-x)}}$ to decide how likely we are to turn on a unit. This function prefers activations ${\bf W}_i{\bf x}$ which are larger than their learned threshold ${\bf b}_i$. Our empirical evaluation invariably found better results with deeper functions. Deeper functions allow the gating units to model richer interactions between the factors. Therefore we propose a sigmoid multi-layer network
\begin{equation}
g({\bf W}{\bf x}) = \sigma({\bf W}^{(2)}\sigma({\bf W}^{(1)}{\bf W}{\bf x} + {\bf b}^{(1)}) + {\bf b}^{(2)}) \nonumber
\end{equation}
with the parameters $\{{\bf W}^{(1)}, {\bf b}^{(1)}, {\bf W}^{(2)}, {\bf b}^{(2)}\}$. This equation showcases a single 
hidden layer but models with additional layers can also be considered.  Defining $g$ as a neural network allows joint training with $f$ through gradient descent.

The difficulty for training resides in computing gradients through  sampling operations ${{\bf g} \sim \mathcal{B}(p=g({\bf W}{\bf x}))}$ which makes ${\bf h}$ binary. \cite{bengio2013estimating} proposed a solution based on reinforcement learning, while \cite{tang2013learning} explored a variational learning approach. These solutions unfortunately results in high-variance gradient estimates. We use a lower variance estimator introduced recently by \cite{raiko2014techniques}. This approach decomposes the stochastic units into
\begin{equation}
{\bf g}_i = g_i({\bf W}{\bf x}) + \epsilon_i
\quad\textrm{with}\quad
\epsilon_i = \begin{cases}
    1 - g_i({\bf W}{\bf x})       & \quad \text{with probability } g_i({\bf W}{\bf x})\\
    - g_i({\bf W}{\bf x})  & \quad \text{with probability } 1 - g_i({\bf W}{\bf x})\\
  \end{cases}\nonumber
\end{equation}
which expresses the Bernoulli unit as the sum of the deterministic term $g_i({\bf W}{\bf x})$ and the stochastic term $\epsilon$. The strategy propagates the gradient only through the deterministic term which is the output of the gating function and ignores the gradient coming from $\epsilon_i$. Noting that the term $\epsilon$ has zero mean, that is $E[{\bf g}]=g({\bf W}{\bf x})$, \cite{raiko2014techniques} finds this method incurs only a small bias and works well in practice. This strategy is simple to implement as it amounts to sampling the probabilities ${\bf g}_i$ for the forward pass, and back-propagating through the gating function as if there was no sampling. 

\section{Deep Linearizing Belief Nets}

Multiple layers of non-deterministic variables ${\bf g}^{(1)}, \ldots, {\bf g}^{(L)}$ can be beneficial. This factorizes the latent variable distribution as
$
p({\bf g}|{\bf x}) = \prod_l^L p({\bf g}^{(l)}|{\bf x}, {\bf g}^{(l-1)})
$ 
and increases modeling efficiently. The resulting deep LBN is a hierarchical mixture that has layers of shared factors. This extension yields the following density for two layers
\begin{equation}
p({\bf y}|{\bf x}) = \sum_{{\bf g}^{(1)},{\bf g}^{(2)}} p({\bf g}^{(2)}|{\bf x}, {\bf g}^{(1)}) p({\bf g}^{(1)}|{\bf x}) p({\bf y}|{\bf x},{\bf g}^{(1)},{\bf g}^{(2)}) \nonumber.
\end{equation}
In that case, the linear expert adds a new linear layer along with the new gating units  
$$
p({\bf y}|{\bf x}) \propto e^{-\frac{1}{2}\|f({\bf x}, {\bf g}^{(i)})-y\|^2}
\quad
\textrm{with}
\quad
f({\bf x}, {\bf g}^{(1)}, {\bf g}^{(2)}) 
= 
{\bf U}({\bf g}^{(2)}\otimes{\bf V}({\bf g}^{(1)}\otimes{\bf W}{\bf x})).
$$
The distribution of the first layer gating units remain unchanged while the gating units of the second layer follow 
$
p({\bf g}^{(2)}|{\bf x}, {\bf g}^{(1)}) 
= 
\mathcal{B}(p=g^{(2)}({\bf V}({\bf g}^{(1)}\otimes{\bf W}{\bf x}))).
$
One can note that the second layer gating function takes the activation of the second layer linear units as input.

\section{Related Work}
\label{sec:related_work}

\cite{neal1992connectionist} proposed one of the earliest uses of neural networks for modeling multi-modal distributions. This model is not suitable for continuous distributions because it is piece-wise constant $p({\bf y}|{\bf x},{\bf g}) = p({\bf y}|{\bf g})$ - it does not vary smoothly with the input. \cite{bishop1994mixture} described a model more suitable for continuous problems called mixture density networks (MDN). The approach forgoes stochastic latent variables and instead has the network directly predict the means of $K$ Gaussians. While it can model continuous distributions, the model is intractable in many cases because the number of parameters grows linearly with the number of modes.

\cite{tang2013learning} improved upon the SBN with the addition of deterministic hidden variables to model continuous distributions. The stochastic and deterministic units are concatenated at each layer to form the representation. In effect, the contributions of the deterministic and stochastic units are combined additively at the next layer. Thus the stochastic units cannot easily switch on or off the deterministic factors. Moreover, training the network can be cumbersome. It requires training a deep deterministic neural network at the same as the stochastic units through a high variance variational bound. By contrast, the deterministic part of LBNs is linear and easy to train because it is linear. Moreover, we optimize likelihood directly through a technique with lesser variance, see Section~\ref{sec:implementation}. The difficulty of training SFNNs is discussed by \cite{raiko2014techniques}. This work finds that better performance can be achieved by training only the deterministic units and setting the probability of activation of the Bernoulli
units to $p=0.5$. The random units are like additional inputs to the network but it differs from simply injecting noise because the mixture loss of Equation \ref{eqn:mixture} is used. We refer to this method as randomized SFNN (RSFNN) from hereon. The challenge here is that this method is less efficient than adapting the stochastic units. More generally, these different approaches relies on the additive combination of the deterministic units with the stochastic units. The estimates of the gradients of the weights of these units have different variance -- with much higher variance for the stochastic part. Training can get trapped in configuration that does not fully exploit the stochastic units. \cite{DBLP:journals/corr/GoroshinML15} explores an alternative strategy and forgo full probabilistic modeling to focus on MAP inference. It introduces non-deterministic hidden variables and performing MAP inference on them. MAP inference at training time can be intensive computationally. 

The linearizing net has connections to the spike and slab RBM \citep{courville2011spike} which has both Gaussian and Bernoulli units that interact multiplicatively. However, the spike and slab RBM is more difficult to train because it requires MCMC. The architectures are also different because  spike and slab places the binary units before the linear units in the flow graph which would make gradient descent challenging.


\section{Experiments}\label{sec:experiments}

This section evaluates the modeling power of LBNs and other stochastic networks on multi-modal distributions. In particular, we  will experimentally confirm the claim that LBNs learn faster and generalize better than other stochastic networks described in the literature. To do so, we consider the tasks of modeling facial expressions and image denoising on benchmark datasets. We train networks with the Adam \citep{KingmaB14} gradient-based optimizer and the parameter initialization of \citep{glorot2010understanding}. We found it was optimal to initialize the biases of all units in the gating networks to $-2$ to promote sparsity. The hyper-parameters of the network are cross-validated using a grid search where the learning rate is always taken from $\{10^{-3}, 10^{-4}, 10^{-5}\}$, while the other hyper-paremeters are found in a task specific manner. All experiments are run the same hardware (Nvidia Tesla K40m GPUs) and all compared techniques are given the same training time.

\subsection{Modeling facial expressions}

\begin{figure}[t]
\centering
\includegraphics[width=0.9\textwidth]{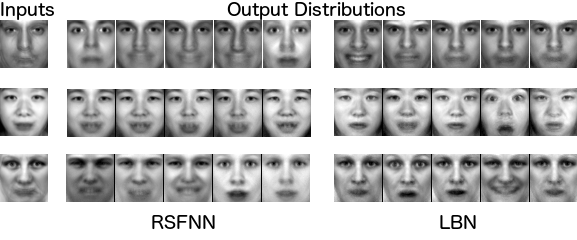}
\caption{Output distributions for 3 input test subjects predicted by the RSFNN and LBN. The LBN generates a distribution of faces that have more varied expressions and that maintain the identity more closely. It also showcases the ability of the network to output a diverse distribution of images instead of just the mean prediction.\label{fig:faces_comparison}}
\end{figure}

This section compares different approaches to stochastic feed-forward neural networks on the task of predicting the distribution of facial expressions of a person given a picture with a neutral expression as in \cite{tang2013learning} and \cite{raiko2014techniques}. The input ${\bf x}$ is the average face of the person and we have a distribution of pictures ${\bf y}^\text{sad}, \ldots, {\bf y}^\text{angry}$ of that person with 7 different emotions. The goal is to be able to produce the full set of facial expressions for a face we have not seen before. The pictures are taken from the Toronto Face Dataset (TFD) \citep{susskind2010toronto} which contains 4,000 images of 900 subjects which were asked to display 7 different emotions. Following the setting of \cite{tang2013learning}, we randomly selected 95 subjects with 1,318 images for training, 5 subjects with 68 images for validation and 24 individuals totaling 343 images were used as a test set. This reproduces the setting from \cite{tang2013learning} as closely as possible. The networks were trained for 200 iterations on the training set with up to $k=200$ Monte Carlo samples to estimate the expectation over outcomes.

We consider various methods including RSFNNs~\cite{raiko2014techniques}, mixtures of factor analysers (MFA), conditional Gaussian RBMs (C-GRBM) and SFNNs. The stochastic networks are trained with 4 layers with either 128 or 256 deterministic hidden units. ReLU activations are used for the deterministic units as they were found to be good for continuous problems. The 2 intermediary layers are augmented with either 32 or 64 random Bernoulli units. The number of hidden units in the LBNs was chosen from $\{128, 256\}$ with the number of hidden layers fixed to 1. The gating network has 2 hidden layers with $\{64, 128\}$ hidden units. The hidden units of the gating network are also sampled under a Bernoulli distribution. This allows the gating pattern ${\bf h}$ to be itself to be multi-modal and results in better results.

\begin{table}[ht]
\centering
\begin{tabular}{c | c |c | c | c | c | c | c | c}
   MFA & MDN & C-GRBM & SFNN & RSFNN & LBN & LBN  & LBN \\
    &  &  &  & ($K=200$) & ($K=1$) & ($K=50$) & ($K=200$) \\
  \hline
1406 & 1321 & 1146 & 1534 & 2250 & 2007 & 2564 & {\bf 2691} \\
\end{tabular}
\caption{Average test log-likelihood predicting facial expressions on TFD. The LBN improves with the number of Monte Carlo samples and reaches state-of-the-art results at $K>50$.\label{tbl:tfd}}
\end{table}

Table \ref{tbl:tfd} reports our results for RSFNN and LBN as well as the results from~\cite{tang2013learning} for the other techniques. Test likelihood is evaluated through Monte-Carlo sampling, like for training. The significant difference between LBNs and RSFNNs compared to the other models can be explained by their use of training methods which have much more variance. The C-GRBM requires Gibbs sampling during training for instance. The results
shows superior generalization performance for the LBN with a difference of 400 nats with RSFNNs. We also find that LBN networks converge faster than the RSFNNs in terms of training epochs. By looking at the predicted facial expressions in Figure \ref{fig:faces_comparison}, we confirm that the model is able to output a rich distribution of images. We also find that the gap in performance can be explained by the fact that the RSFNNs have difficulty maintaining the identity of the face. In the RSFNN, the identity can only be encoded by the deterministic units so this shows that the RSFNN has difficulty conditioning on that information. These issues are resolved by the LBN in two ways: the stochastic units can also learn to condition on identity and the multiplicative interactions ease making tie-breaking choices. Interestingly, the log-likelihood on the training set for the LBN is 3081 nats while the RSFNN reaches 2938 nats. Thus the RSFNN allows memorizing the training set in a similar manner, but does not lead to conditional models that generalize as well. 


\subsection{Image Denoising}

Denoising is a common and challenging task to evaluate unsupervised models of natural images. Typically, the goal is to learn to remove homogeneous additive Gaussian noise from the input. The noise destroys information and so the model must infer the original image from the corrupted signal. In order to do this, the model must discover local statistics of the distribution of images to map a corrupted image to the nearest valid image. \cite{bengio2013generalized} showed that under mild conditions the denoising models learn the transition operator of a Markov chain whose stationary distribution is the data distribution. Denoising is an interesting application for LBNs because it is an inverse problem. There is a distribution of clean images that may correspond to a corrupted image, and this distribution may be multimodal for highly corrupted images. This occurs when the noise destroys information beyond perfect recovery. Previous approaches such as the state-of-the-art BM3D~\citep{dabov2009bm3d} method ignore this difficulty and simply predict the conditional average. This strategy does not predict plausible images when high noise has truly destroyed information since the conditional average exhibits statistics far from genuine images in that case. Typically, such methods results in blurry predictions for highly corrupted images. On the other hand, this problem can be addressed by a model that predicts a distribution from which a set of plausible images explaining the noisy one can be sampled. Of course, this is a difficult problem and we explore here a step in that direction. We show that the LBN improves upon the conditional SBN for image patches and we compare LBNs to state-of-the art methods on full images in the last section.

We report results using the Peak Signal to Noise Ration (PSNR) which is given by $PSNR({\bf x}, {\bf y}) = -10\log(\|{\bf x} - {\bf y}\|^2/N)$ where ${\bf x}, {\bf y} \in [0, 1]^N$. The PSNR is an approximation to the human perception of denoising quality and it relates to the distance between corrupted and clean image. It does not operate on distributions but we use it here since it a well accepted state-of-the-art measure. It would have been better to have a distributional measure but finding a better denoising quality measure is an open problem. Nonetheless, we can compute the PSNR with a representative point from the distribution either by drawing a random sample, computing the mean or the MAP - which would be the gold standard. For one hidden layer models, we compute the mean exactly by setting ${\bf g} = g({\bf x})$. We find that using this strategy works well in practice. For deeper models, we simply draw a sample from the model since the other quantities are difficult to estimate exactly. In addition to reporting PSNR, we also qualitatively evaluate the diversity of the distribution visually.

\subsubsection{Image patches}\label{sec:patches}

\begin{figure}[ht]
\centering
\includegraphics[width=0.6\textwidth]{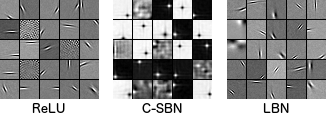}
\caption{Filters learned by denoising $19\times 19$ image patches on Imagenet for 2 layer models. The C-SBN learns poor global features while the ReLU and LBN naturally learn Gabor-like filters.\label{fig:filter_comparison}}
\end{figure}

In this section, we will show that LBNs significantly improves upon conditional SBNs for denoising image patches. We consider small image patches here for computational reasons, as training on larger patches is computationally intensive. We extract $19\times 19$ image patches from the Imagenet dataset. The dataset has close to 1.3M images and we extract a random subsample of 13M grayscale patches - with 10,000 patches held out as a test set. We will consider the problem of denoising with a Gaussian with standard deviation 25 (over 255). We preprocess the patches by reducing to the $[0, 1]$ interval, substracting the mean of $\mu=0.5$ and dividing by the standard deviation $\sigma=0.2$.

We will compare conditional SBNs, LBNs and deterministic ReLU networks as a good general baseline. The LBNs have 1 non-deterministic hidden layer and the gating function has 3 layers of sigmoids. The ReLU networks were trained with either 2 or 6 layers, with 6 layers always improving results. We trained the C-SBNs with 6 layers also. All models were trained with 1024 hidden units. The 6 layer ReLU network and C-SBN have close to 4.9M parameters, while the LBN has 1M less with 3.8M parameters. The gradient of the binary units in the C-SBNs is found using the same estimator used in the LBNs so they are directly comparable. The Monte Carlo estimation of the expectation during training is computed with either 1 or 10 samples. All networks are trained with 10 iterations over the 13M patches.

\begin{figure}[ht]
\centering
\includegraphics[width=0.45\textwidth]{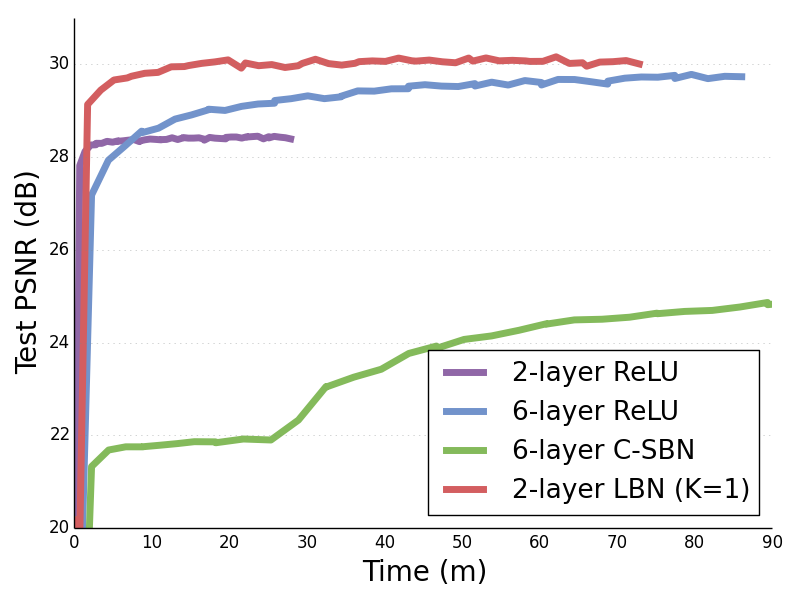}
\caption{Average test PSNR versus training time for patch-wise image denoising. LBNs achieve superior results with a smaller budget of parameters and time compared to conditional SBNs and even a deep ReLU network.\label{fig:denoising_curve}}
\end{figure}

We have plotted the learning curves with respect to training time in Figure \ref{fig:denoising_curve}. The LBN significantly outperforms the conditional SBN in terms of training time and test PSNR. The improved convergence speed can be explained in part by the linear units which allow the gradient to flow better through the network. What's more, the fact that LBNs allow higher PSNRs showcases the superior approximation efficiency of a mixture of linear models over a mixture of constants. Surprisingly, the LBN compares favorably in terms of converge speed even compared to deterministic ReLU networks. The power of the representation learned by LBNs is evidenced by the fact that it requires 6 layers of ReLUs to come close to the same level of performance. Unlike with the ReLU representations, the gating units can resolve ambiguity and competition in the representation. On the qualitative side, we can see ReLU and LBN networks learn localized Gabor filters while the C-SBN learns point detectors. These results show that the LBN model successfully addresses some of the flaws of conditional SBNs (see Appendix \ref{sec:more_patches} for more).

\subsubsection{Full images}\label{sec:full_images}

In this section, we evaluate the problem of training LBNs with multiple layers of non-deterministic variables on denoising full images. The state-of-the art method for denoising natural images is BM3D \citep{dabov2009bm3d}. Neural methods \citep{burger2012image} were state-of-the-art but they were surpassed by later versions of BM3D. BM3D is a non-parametric denoising algorithm that uses self-similarity between patches to average out the noise. While we can expect this method to work well for small noise distributions, it produces blurry images for high noise distributions.

In order to scale to large images we use a convolutional architecture for the LBNs. In practice, this amounts to replacing the dot products by convolutions - both for the activations of the linear units and the gaters. We found that using 128 convolutional kernels of size $9\times 9$ for the linear and gating units produced good results. The network has 4 convolutional hidden layers in total and 3 layer gating functions for LBNs. We do not use any spatial pooling because the loss of information would be detrimental to denoising. The output pixels have a scalar bias so as to not constrain the size of the images that can be generated. We extract 1M $64\times64$ image patches from Imagenet as our training set for highly corrupted images and 6M $29\times29$ patches otherwise. We use these large image patches instead of the full image to save computation time during training but the same model can be applied to large images at test time. The networks can take up to a week of computation time to train and so in the interest of the PSNR evaluation we set the number of Monte Carlo samples to $K=1$. We evaluate the quality of the distribution with a model trained with $K=10$. We compare our algorithm
to Gaussian Scale Mixture (GSM) \citep{portilla2003image}, Field of Experts (FoE) \citep{roth2005fields}, K-SVD \citep{elad2006image}, BM3D \citep{dabov2009bm3d} and Learned Simultaneous Sparse Coding (LSSC) \citep{mairal2010online} on the standard 11 test images they used.
\begin{table}[h]
\centering
\begin{tabular}{ l || c | c | c |c | c | c }
  Image & GSM & FoE & KSVD & BM3D & LSSC & LBN \\
  \hline \hline
 Barbara & 22.61 & 19.77 & 21.89 & {\bf 23.62} & 23.59 & 23.22 \\
 Boat    & 23.75 & 20.80 & 22.81 & 23.97 & 23.84 & {\bf 24.62} \\
 C.man   & - & - & - & 23.07 & 23.08 & {\bf 24.08} \\
 Couple  & - & - & - & 23.51 & 23.28 & {\bf 24.27} \\
 F.print & 21.22 & - & 18.30 & {\bf 21.61} & 21.26 & 21.52 \\
 Hill    & - & - & - & 24.58 & 24.44 & {\bf 25.15} \\
 House   & 25.11 & 21.66 & 23.71 & 25.87 & 25.83 & {\bf 27.21} \\
 Lena    & 25.64 & 21.87 & 25.64 & 25.95 & 25.82 & {\bf 26.75} \\
 Man     & - & - & - & 24.22 & 24.00 & {\bf 24.76} \\
 Montage & - & - & - & 23.89 & - & {\bf 25.45} \\
 Peppers & 22.60 & 19.60 & 21.75 & 23.39 & 23.00 & {\bf 24.13}\\
\end{tabular}
\caption{Denoising PSNR on standard test images with $\sigma=100$. The LBN reaches state-of-the-art.
\label{tbl:denoising}}
\end{table}

\begin{figure}[t]
\centering
\includegraphics[width=0.4\textwidth]{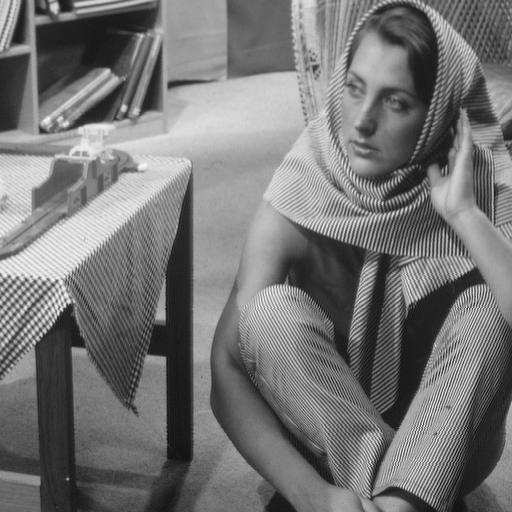}
\includegraphics[width=0.4\textwidth]{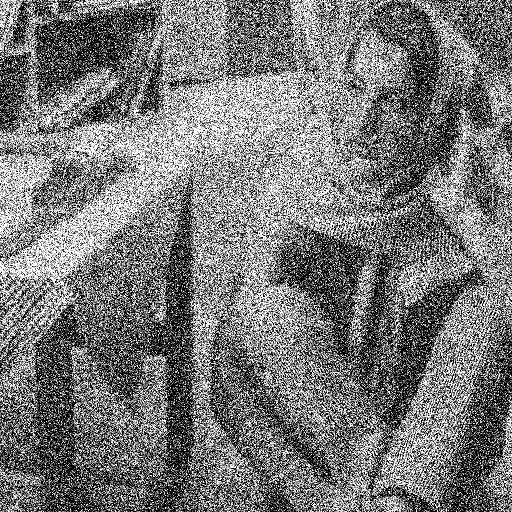}
\includegraphics[width=0.4\textwidth]{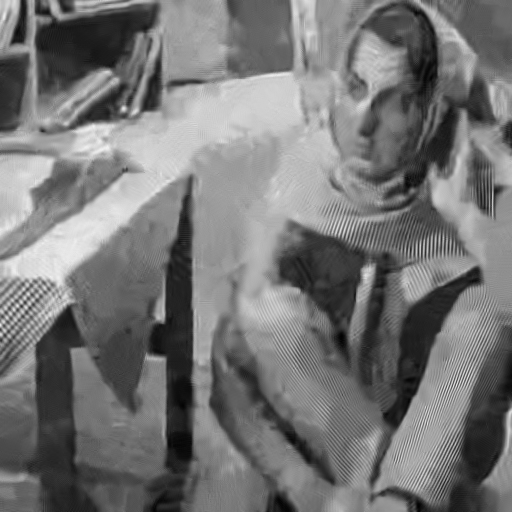}
\includegraphics[width=0.4\textwidth]{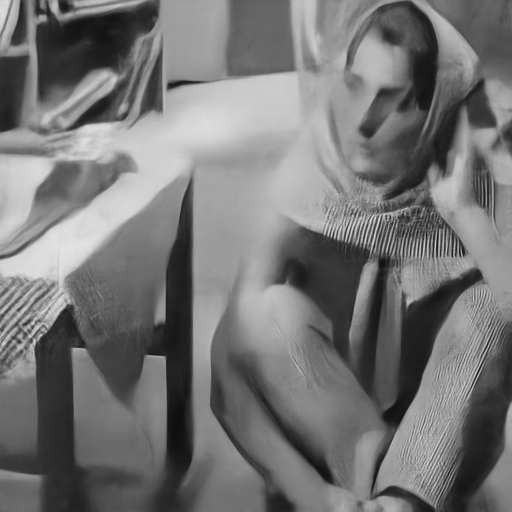}
\caption{Visualization of denoising for the Barbara image with $\sigma=100$. Top-left is the original, top-right the noisy image, bottom-left is BM3D and bottom right is the output of a LBN. The LBN produces qualitatively more pleasing denoised images. There is a video of the distribution at \url{http://ynd.github.io/lbn_denoising_demo/}.\label{fig:barbara}}
\end{figure}

Table \ref{tbl:denoising} shows the results denoising the standard test images in the setting of a Gaussian noise with $\sigma=100$. We find that LBNs overall produce a substantial improvement over the state-of-the-art methods (Appendix \ref{sec:more_denoising} shows state-of-the-art results for other settings). The two images where BM3D sets the state-of-the-art have highly repetitive structure which favors its approach. Figure \ref{fig:barbara} shows that the LBN produces arguably better qualitative results even on those images. In particular the images are much sharper and more closely resemble natural images. In the context of lower noise, the LBN also achieve state-of-the-art results with an average test PSNR of 30.70 dB at $\sigma=25$ (with BM3D at 30.4 dB) but it learns a deterministic mapping. This makes sense because there is little uncertainty over the clean images. At $\sigma=100$, we find through visual inspection that the model can generate a distribution of alternatives (Videos available at \url{http://ynd.github.io/lbn_denoising_demo/}). In these videos we see better variety in the distribution with $K=10$ compared to $K=1$. However, as we can also see in Figure \ref{fig:barbara} the model does not capture enough about the distribution to produce very plausible denoised images - there is still much averaging. This suggests that estimating the expectation over outcomes with Monte Carlo samples is not a good enough estimator for highly multimodal distributions. An interesting direction would be to improve our estimate using importance sampling. We believe this is a good direction to further improve denoising results. Alternatively, one could forgo maximum likelihood training of the model and use an adversarial objective instead \citep{goodfellow2014generative}. Nonetheless - even if the training method has to be improved - our results confirm the LBN model improves upon traditional conditional belief net models.

\section{Conclusion}

This work introduces linearizing belief nets (LBN), a new class of conditional belief network. As a belief network, a LBN relies on stochastic binary units but is well suited to model continuous distributions. Contrary to prior work, LBN stochastic units act as gaters to a deep linear network. This multiplicative interaction between stochastic and deterministic units allows better cooperation between the two parts of the network compared to prior additive strategies. Moreover, LBN linear units propagate continuous information efficiently and combined with stochastic binary gating acts as skip-connections that prevent gradient diffusion and help learning. Our experiments confirm these advantages. Our facial expression generation experiments result in better generalization and faster convergence for LBN compared to alternative belief networks. Our image denoising experiments also report better signal-to-noise ratio than previous work. Overall, this work proposes a generic model that can be relevant to various inverse problems. In the future, we want to investigate alternatives to our current Monte Carlo maximum likelihood training. In particular, we consider adversarial training or importance sampling to model distribution with more modes efficiently. 

\section*{Acknowledgements}
The authors would like to thank Marc'Aurelio Ranzato for insightful comments and discussions.

{
\small
\bibliography{iclr2016_conference}
\bibliographystyle{iclr2016_conference}
}

\newpage
\appendix

\section{Additional results}

\subsection{LBNs on natural image patches (From Section \ref{sec:patches})}\label{sec:more_patches}

\begin{figure}[h]\label{fig:denoising_hist}
\centering
\includegraphics[width=0.4\textwidth]{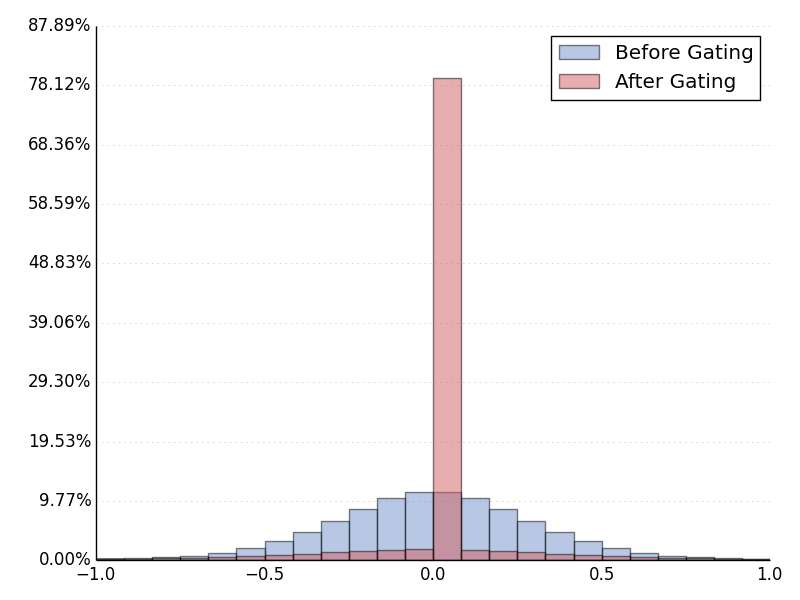}
\caption{Histogram of the hidden activations of the LBN before (blue) and after (red) gating. The gating units resolve ambiguity and remove redundant activations leading to a sparser representation.}
\end{figure}

\begin{figure}[h]\label{fig:full_filters}
\centering
\includegraphics[width=.99\textwidth]{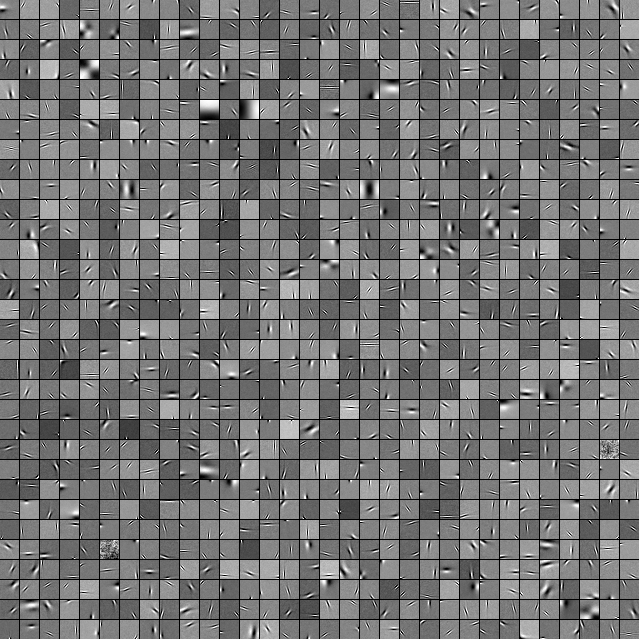}
\caption{Full 1024 feature detectors of the the LBN. Virtually all the filters converge to meaningful feature detectors, which is not the case for several models.}
\end{figure}

\subsection{LBNs on full natural images (From Section \ref{sec:full_images})}\label{sec:more_denoising}

\begin{table}[h]
\centering
\begin{tabular}{ l || c | c | c |c | c | c }
  Image & GSM & FoE & KSVD & BM3D & LSSC & LBN \\
  \hline \hline
 Barbara & 29.13 & 27.04 & 29.60 & {\bf 30.72} & 30.47 & 30.24 \\
 Boat    & 29.37 & 28.72 & 29.28 & 29.91 & 29.87 & {\bf 30.25} \\
 C.man   & - & - & - & 29.45 & 29.51 & {\bf 29.98} \\
 Couple  & - & - & - & 29.72 & 29.61 & {\bf 30.19} \\
 F.print & 27.45 & - & 18.30 & 27.7 & 27.62 & {\bf 27.77} \\
 Hill    & - & - & - & 29.85 & 29.80 & {\bf 30.06} \\
 House   & 31.40 & 31.11 & 32.15 & 32.86 & 33.15 & {\bf 33.31} \\
 Lena    & 31.69 & 30.82 & 31.32 & 32.08 & 31.87 & {\bf 32.51} \\
 Man     & - & - & - & 29.62 & 29.63 & {\bf 30.05} \\
 Montage & - & - & - & 32.37 & - & {\bf 32.90} \\
 Peppers & 29.21 & 29.20 & 29.73 & 30.16 & 30.21 & {\bf 30.81}\\
\end{tabular}
\label{tbl:denoising_sigma_25}
\caption{Denoising PSNR on standard test images with $\sigma=25$. We compare our algorithm
to Gaussian Scale Mixture (GSM) \citep{portilla2003image}, Field of Experts (FoE) \citep{roth2005fields}, K-SVD \citep{elad2006image}, BM3D \citep{dabov2009bm3d} and Learned Simultaneous Sparse Coding (LSSC) \citep{mairal2010online}.}
\end{table}

\begin{table}[h]
\centering
\begin{tabular}{ l || c | c | c |c | c | c }
  Image & GSM & FoE & KSVD & BM3D & LSSC & LBN \\
  \hline \hline
 Barbara & 23.65 & 23.15 & 25.47 & {\bf 27.23} & 27.06 & 26.37 \\
 Boat    & 24.79 & 24.53 & 25.95 & 26.78 & 26.74 & {\bf 27.25} \\
 C.man   & - & - & - & 26.12 & 26.42 & {\bf 27.11} \\
 Couple  & - & - & - & 26.46 & 26.30 & {\bf 27.02} \\
 F.print & 22.40 & - & 18.30 & {\bf 24.53} & 24.25 & 24.30 \\
 Hill    & - & - & - & 27.19 & 27.05 & {\bf 27.48} \\
 House   & 26.41 & 26.74 & 27.95 & 29.69 & 30.04 & {\bf 30.58} \\
 Lena    & 26.84 & 26.49 & 27.79 & 29.05 & 28.87 & {\bf 29.45} \\
 Man     & - & - & - & 26.81 & 26.69 & {\bf 27.13} \\
 Montage & - & - & - & 27.9 & - & {\bf 29.03} \\
 Peppers & 24.00 & 24.52 & 26.13 & 26.68 & 26.62 & {\bf 27.21}\\
\end{tabular}
\label{tbl:denoising_sigma_50}
\caption{Denoising PSNR on standard test images with $\sigma=50$. We compare our algorithm
to Gaussian Scale Mixture (GSM) \citep{portilla2003image}, Field of Experts (FoE) \citep{roth2005fields}, K-SVD \citep{elad2006image}, BM3D \citep{dabov2009bm3d} and Learned Simultaneous Sparse Coding (LSSC) \citep{mairal2010online}.}
\end{table}

\subsection{Links between LBNs and ReLU networks}\label{sec:relu}

ReLU networks can be seen as a particular deterministic subset of the LBN family of networks. The function of a ReLU network is given by $f({\bf x}) = {\bf V}\text{max}(0, {\bf W}{\bf x}) = {\bf V}(({\bf W}{\bf x} > 0)\circ {\bf W}{\bf x})$. This is the form of a LBN with gating units  sampled from a Dirac delta distribution ${\bf h} \sim \delta(g({\bf W}{\bf x}))$ where the gating function $g({\bf W}{\bf x}) = {\bf W}{\bf x} > 0$. Altogether we have $f({\bf x}) = {\bf V}({\bf h}\circ {\bf W}{\bf x})$ which is the form of a linearizing network.

We can relax the determinism of the gating units by using a Bernoulli distribution with $g({\bf W}{\bf x}) = \sigma({\bf W}{\bf x} + {\bf b})$. This function is very close to that of the ReLU because the sigmoid is a relaxation of the threshold function. Interestingly, by performing MAP inference at test time we recover the ReLU since $\sigma({\bf W}{\bf x} + {\bf b}) > 0.5 = {\bf W}{\bf x} + {\bf b} > 0$. We have found experimentally that this simple model produces similar results to the ReLU. This link gives some intuition as to why the LBN model is more powerful than simple ReLUs. The LBNs can have more powerful gating functions while the gating function of ReLus is fixed and much less powerful (it is a simple threshold).

\end{document}